# Tracking of Fingertips and Centres of Palm using KINECT

Jagdish L. Raheja       Ankit Chaudhary       Kunal Singal

*Abstract—* Hand Gesture is a popular way to interact or control machines and it has been implemented in many applications. The geometry of hand is such that it is hard to construct in virtual environment and control the joints but the functionality and DOF encourage researchers to make a hand like instrument. This paper presents a novel method for fingertips detection and centres of palms detection distinctly for both hands using MS KINECT in 3D from the input image. KINECT facilitates us by providing the depth information of foreground objects. The hands were segmented using the depth vector and centres of palms were detected using distance transformation on inverse image. This result would be used to feed the inputs to the robotic hands to emulate human hands operation.

## I. Introduction

Many researchers have proposed different methods for dynamic hand gesture recognition using fingertip detection dedicated to different applications. It begins a new era to see the existing problems in the area of Hand Gesture recognition, as the sensors which can give depth information, are available. Microsoft KINCT is one of the examples and it is able to detect individual finger motion. There exist several limitations in the past approaches, see [3]. Garg [4] used 3D images in his method to recognize the hand gesture, but this process was complicated and inefficient. The focus should be on efficiency with the accuracy as processing time is a very critical factor in real time applications. Yang [16] analyzed the hand contour to select fingertip candidates, then finds peaks in their spatial distribution and checks local variance to locate fingertips. This method was not invariant to the orientation of the hand. There are other methods, which are using directionally Variant templates to detect fingertips [6][12]. Few other methods are dependent on specialized instruments and setup, like the use of infrared cameras [6], stereo cameras [17], a fixed background [2][8] or use of markers on hand. Raheja [11] showed an efficient real time technique for natural hand with simple background, where orientation constraint was covered.

This paper describes a novel method of fingertips and centre of palms detection in dynamic hand gestures generated by either one or both hands without using any kind of sensor or marker. We call it Natural Computing as no sensor, marker or color is used on hands to segment skin in the images and hence user would be able to do operations with natural hand. The detection of moving fingertips in real time video needs a fast and robust implementation of method. Many fingertips detection methods are based on hand Segmentation techniques because it decreases pixel area which is going to process by the algorithm by selecting only areas of interest. However most hand segmentation methods can't do a clearly hand segmentation under some natural conditions like fast hands motion, cluttered background, poor light condition. Poor hand segmentation performance usually invalidates fingertips detection methods. Researchers [6][7][13] used infrared cameras to get a reliable segmentation. Few researchers [2][5][8][9][14][15] limited the degree of the cluttered background, finger's motion speed or light conditions to get a reliable segmentation in their work.

Some of fingertips detection methods can't localize accurately multidirectional fingertips. Researchers [1][2][8][14] assumed that the hand is always pointing upward to get precise localization. In our previous approaches [18][20], we detected fingertips and calculated bended fingers angles in 2D for one hand. This method is more robust and reliable as it is based on depth information given by the KINECT, while in 2D it was based on segmentation methods.

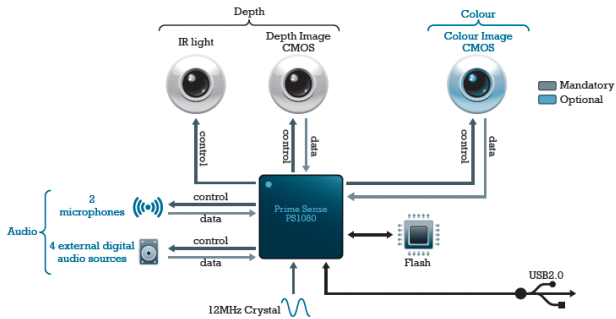

Figure 1. MS KINECT Architecture [10]

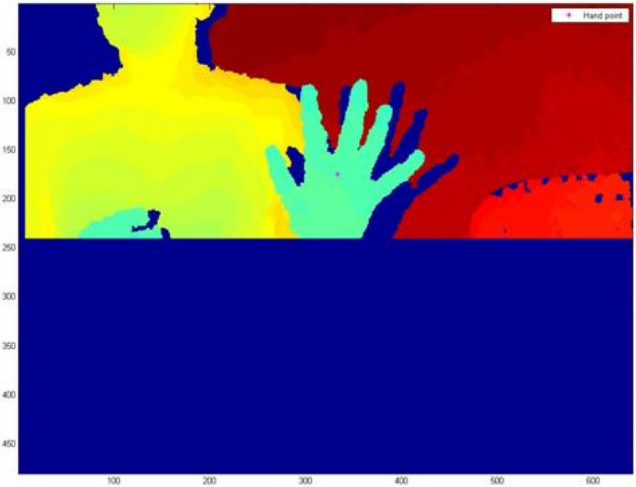

Figure 2. Depth image acquired using KINECT

## II. FINGERTIPS DETECTION IN 3D

The approach of fingertips detection is divided in following steps.

### A. Getting the Depth Image from KINECT

The internal architecture of MS KINECT is shown in Figure 1. It have infrared camera and PrimeSense sensor to compute the depth of the object while the RGB camera is used to capture the images. As Frati [19] stated "It has a webcam–like structure and allows users to control and interact with a virtual world through a natural user interface, using gestures, spoken commands or presented objects and images", it is clear that KINECT is a robust device and could be used in different complex applications. The depth images and RGB image of the object could be getting at the same time. This 3D scanner system called *Light Coding* which employs a variant of image-based 3D reconstruction. The depth output of KINECT is of 11 bit with 2048 levels of sensitivity [21]. The depth value $d_{raw}$ of a point in 3D can be defined as calibration procedure [19]

$$d = K\,tang(Hd_{raw} + L) - O$$

where d is the depth of that point in cm, H is $3.5 \times 10^{-4}$ rad, K=12.36 cm , L = 1.18 rad and O=3.7 cm.

### B. Hands Tracking and detecting Hand Point

Hands tracking and points detection were done using NITE modules, which uses Bayesian Object Localization for hand detection. We used OpenNI modules which provide C++ based APIs for hand detection and tracking. The image with depth information is shown in Figure 2.

### C. Segmentation by Depth

The depth image was segmented after putting a calculated threshold on depth of hand points and the hands were detected by choosing the blob which contains hand point as shown in figure 3.

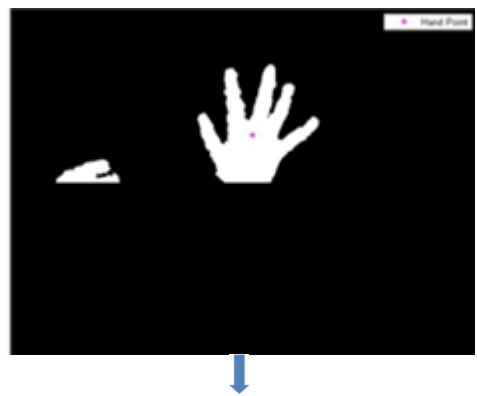

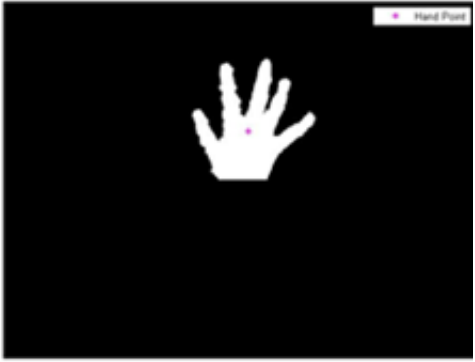

Figure 3. (a) Threshold image, (b) Image of one hand

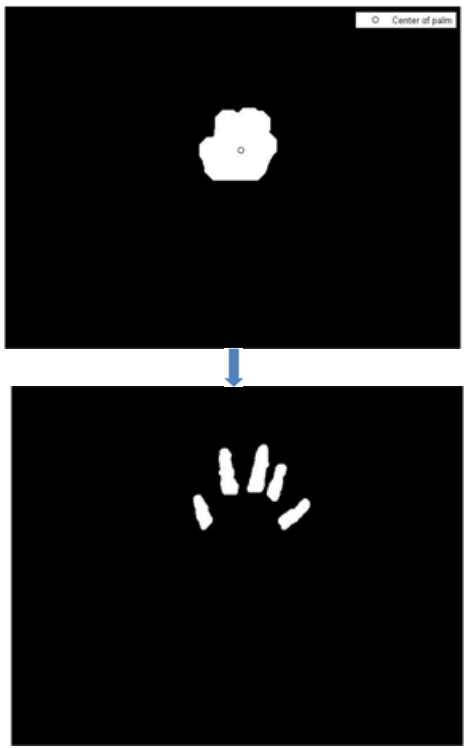

Figure 4. (a) Palm in one hand image, (b) Fingers mask for one hand

### D. Fingertips detections

Fingertips were detected by first finding the palm of the hands which were obtained by applying a big circular filter on the image, so that all the fingers in the images would be removed. Now palm of the hand would be subtracted from the original hand image to get the segmented finger masks as shown in figure 4 and figure 5. The finger masks were multiplied with original depth image to get the depth map of fingers. After examining the depth map, it could be easily find out that at fingertips the value of depth was minimum, it implies that they are closest to camera then remaining objects. So, the fingertips would be detected by finding the minimum depth in each finger. This process would work for both hands also where it applied simultaneously on both hands and results were very encouraging. The results of fingertips detection in real-time for single hand and both hands are shown in figure 6 and figure 9. The orientation of hands would not be a constraint in this approach, as 3D sensor is able to detect it in any direction.

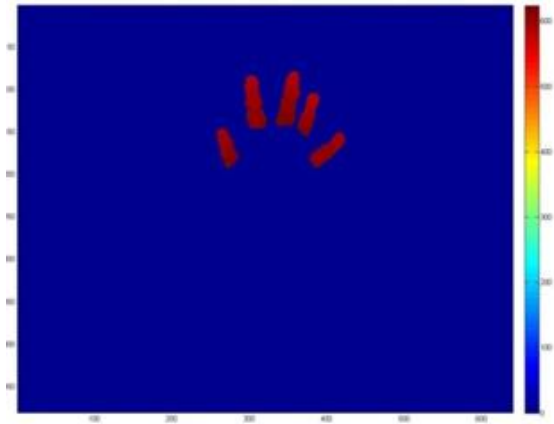

Figure 5. Segmented fingers in depth image

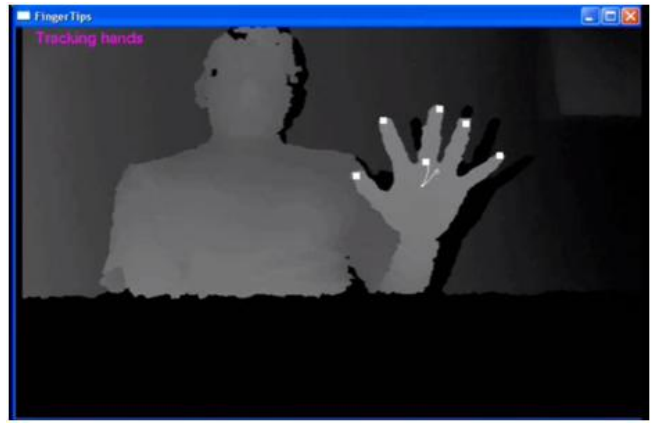

Figure 6. Result of fingertips detection in real time

### III. CENTRE OF PALM DETECTION

Centres of the palms were detected by applying the distance transform on the inverted binary images of hand as shown in figure 7. It was clearly visible that the maximum of the distance transform was giving the centre of the palm on one segment. As this experiment could be used for both hands also, we made clear distinctness to recognize both the hands. As shown in figure 9, if it is single hand the fingertips and centre of palm would be detected in white color while if both hands are detected in the image, the right hand would be detected as white and the left hand would be detected as pink. This color difference makes sure that centre of palm of one hand should not match with the other hand details.

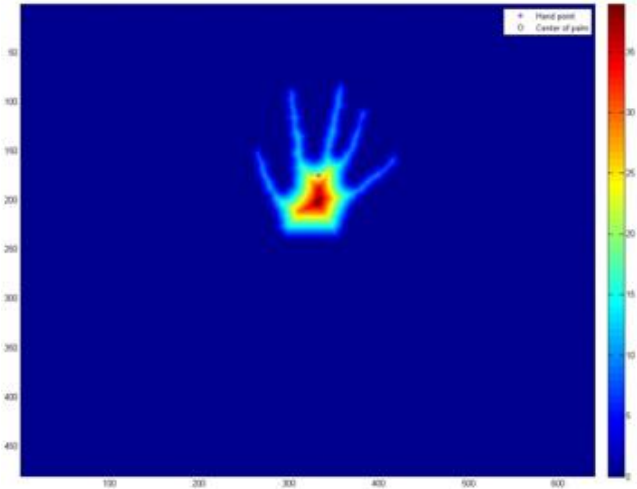

Figure 7. Distance transform of hand

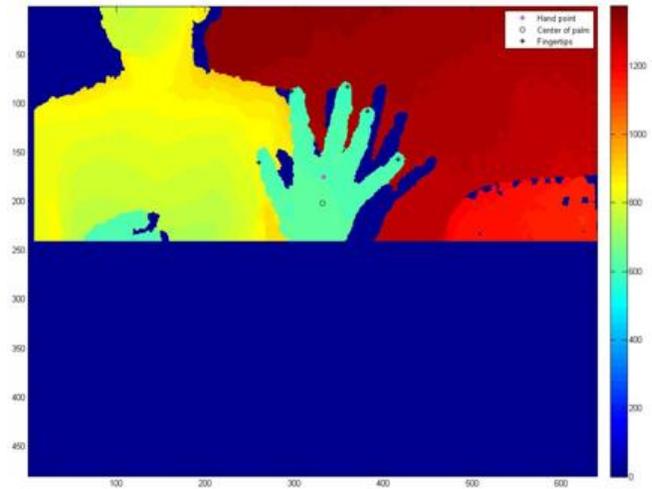

Figure 8. Final result showing hand point, centre of palm, and fingertips

## IV. RESULTS

From our lab setup, we were able to identify fingertips and centre of palm very accurately, even when the fingers were bent by large angle. The accuracy for fingertips detection, when all finger were open was near to 100% while in fully bended fingers case sometimes it was confused. In the case of centre of palm detection, the results were around 90% correct. The whole system was implemented in real time and the results were very encouraging as shown in figure 8 and 9.

## V. CONCLUSIONS

The detection of fingertips and centres of palm has been discussed which will be used in our project 'Controlling the robotic hand using hand gesture'. In this approach the user can show one hand at a time or both hands. This project is still in progress and it will make a significant change in applications which can be harmful to human life. The Movement of user's finger will control the robotic hand by moving hand in front of camera without wearing any gloves or markers.

ACKNOWLEDGMENT

This research was being carried out at Machine Vision Lab, Central Electronics Engineering Research Institute (CEERI/CSIR), A Govt. of India Laboratory, Pilani, INDIA. This research was a part of our current project "Controlling the electro-mechanical hand using hand gesture". Authors would like to thank The Director, CEERI Pilani for providing research facilities and for his active encouragement and support.

**Cite this paper as:**

J.L. Raheja, A. Chaudhary, K. Singal, "*Tracking of Fingertips and Centre of Palm using KINECT* ", In proceedings of the 3rd IEEE International Conference on Computational Intelligence, Modelling and Simulation, Malaysia, 20-22 Sep, 2011, pp. 248-252.

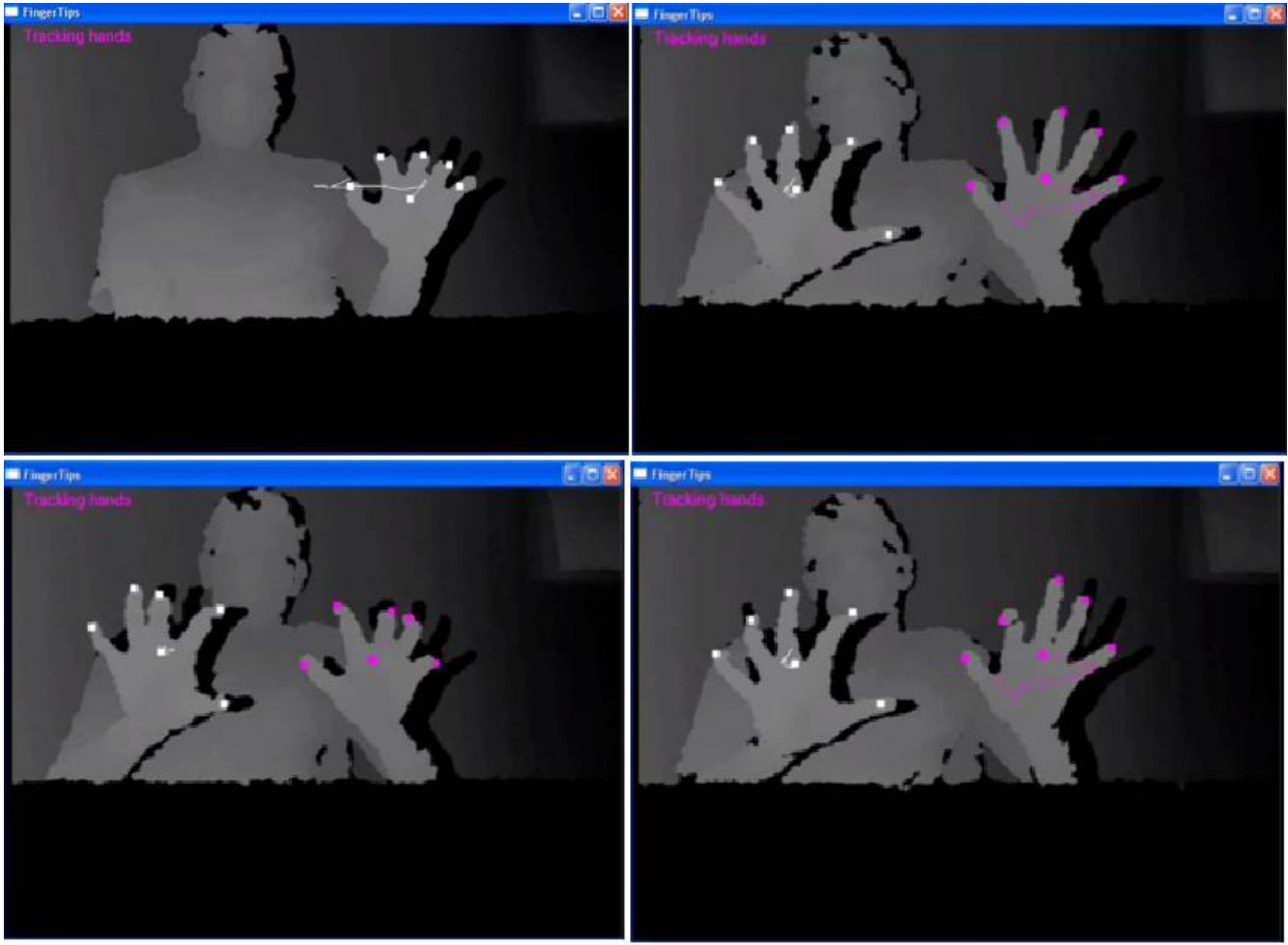

Figure 9. Results of fingertip detection in real time